%% file: emnlp2020.tex
\documentclass[11pt,a4paper]{article}
\usepackage[hyperref]{emnlp2020}
\usepackage{times}
\usepackage{latexsym}
\usepackage{multirow}
\usepackage{graphicx}
\usepackage{pgfplots}
\usepackage{rotating}
\usepackage{floatrow}
\usepackage{afterpage}
\usepackage{amsmath,amssymb}
\usepackage[font={small}]{caption}
\usepackage{algorithm}
\usepackage[noend]{algpseudocode}
\usepackage{url}

\algrenewcommand\algorithmicrequire{\textbf{Input:}}
\algrenewcommand\algorithmicensure{\textbf{Output:}}
\pgfplotsset{compat=1.14}

\usepackage{microtype}

\aclfinalcopy 

\newcommand\csqa{\textsc{CSQA}}
\newcommand\qasc{\textsc{QASC}}

\input macros
\input std-macros

\title{Explaining Question Answering Models through Text Generation}

\author{Veronica Latcinnik$^{1}$ ~~~~~
Jonathan Berant$^{1,2}$ \\
\mbox{}\\
$^1$School of Computer Science, Tel-Aviv University \\
$^2$Allen Institute for AI \\
\small{\texttt{\{veronical@mail,joberant@cs\}.tau.ac.il}}}

\date{}

\begin{document}
\maketitle
\begin{abstract}
\input 00_abstract 
\end{abstract}

\input 01_introduction

\input 02_background
\input 03_model
\input 05_experiments
\input 06_related
\input 07_conclusion

\section*{Acknowledgements}
We thank Inbar Oren and Guy Tevet for their useful suggestions. This research was partially supported by The Israel Science Foundation
grant 942/16, The Yandex Initiative
for Machine Learning and the European Research
Council (ERC) under the European Union Horizons 2020 research and innovation programme
(grant ERC DELPHI 802800).
\newpage
\input 08_rotated_tables
\newpage

\bibliography{all}
\bibliographystyle{acl_natbib}

\onecolumn
\appendix
\input 09_appendix

\end{document}

%% file: macros.tex
\newcommand\nl[1]{{\it``#1''}} 
\newcommand\classifier{p_\text{QA}}
\newcommand\weakclassifier{p_\text{QA}^{\text{sim}}}
\newcommand\strongclassifier{p_\text{QA}^{\text{LM}}}
\newcommand\etoebaseline{\textsc{End2End}}
\newcommand\nointeraction{\textsc{NoInteraction}}
\newcommand\weaksup{\textsc{NoSimClassifier}}
\newcommand\supgen{\textsc{SupGen}}

%% file: std-macros.tex
\newcommand\sA{\ensuremath{\mathcal{A}}}

\newcommand\sW{\ensuremath{\mathcal{W}}}

%% file: 00_abstract.tex
Large pre-trained language models (LMs) have been shown to perform surprisingly well when fine-tuned on tasks that require commonsense and world knowledge. However, in end-to-end architectures, it is difficult to explain what is the knowledge in the LM that allows it to make a correct prediction.
In this work, we propose a model for multi-choice question answering, where a LM-based generator generates a \emph{textual hypothesis} that is later used by a classifier to answer the question. The hypothesis provides a window into the information used by the fine-tuned LM that can be inspected by humans. A key challenge in this setup is how to constrain the model to generate hypotheses that are meaningful to humans. We tackle this by (a) joint training with a simple similarity classifier that encourages meaningful hypotheses, and (b) by adding loss functions that encourage natural text without repetitions. We show on several tasks that our model reaches performance that is comparable to end-to-end architectures, while producing hypotheses that elucidate the knowledge used by the LM for answering the question.

%% file: 01_introduction.tex
\section{Introduction}
\label{introduction}

Language Models (LMs), trained on large amounts of data using self-supervised learning \cite{peters2018deep, devlin2018bert, liu2019roberta, yang2019xlnet, raffel2019exploring}, have been recently shown to encode substantial amounts of knowledge in their parameters \cite{petroni2019language, jiang2019know, talmor2019olmpics}. This has been demonstrated by their ability to answer questions that require common sense and world knowledge, without retrieving information from an external source \cite{trinh2018simple, zhou2019evaluating, roberts2020knowledge, ling2020learning}. For example, the current top model for \textsc{CommonsenseQA} (\csqa{}) \cite{talmor2018commonsenseqa},\footnote{\url{https://www.tau-nlp.org/csqa-leaderboard}} a benchmark testing the ability to answer commonsense questions, answers questions using \textsc{ALBERT} \cite{lan2019albert} only.

\begin{figure}[t]
 \includegraphics[width=1\columnwidth]{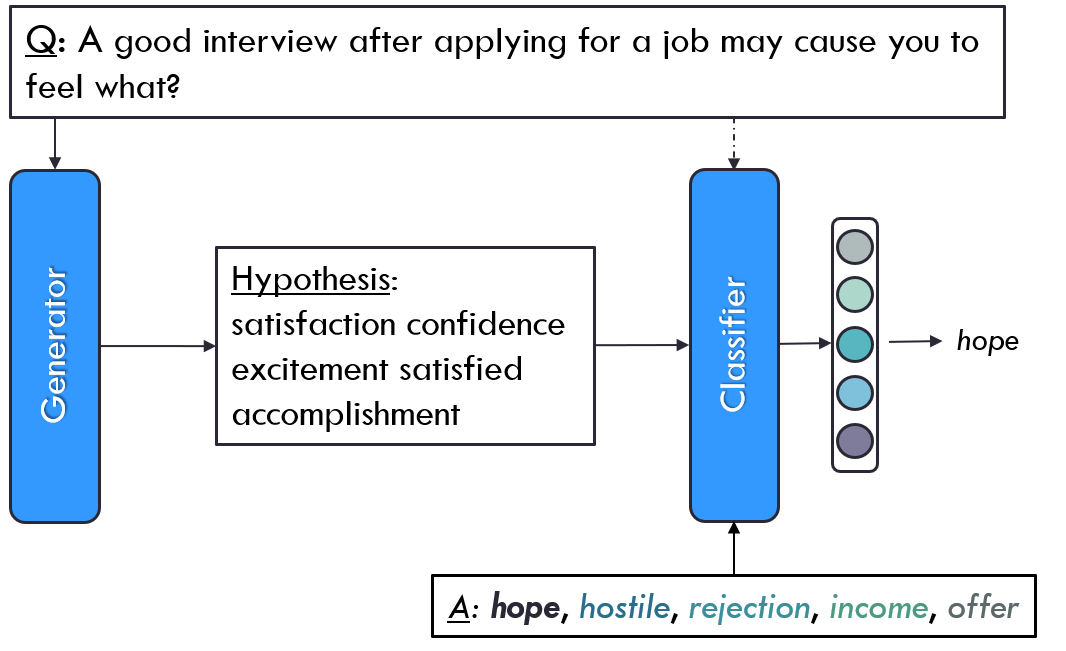}
 \caption{An overview of our approach: a generator takes a question and outputs a textual hypothesis that is used by a classifier to select the answer (example from \csqa{}).
 }~\label{fig:overview}
\end{figure}

Despite these impressive results, most current models are based on end-to-end architectures, where it is difficult to know how the model reached its prediction. In fact, it has been repeatedly shown that often models obtain high performance through ``shortcuts" rather than language understanding \cite{tsuchiya2018performance, poliak2018hypothesis, gururangan2018annotation, geva2019modeling}. This has sparked interest in \emph{explainable models}, where intermediate parts of the model can be inspected and interpreted by humans \cite{lipton2016interpretability,ribeiro2016i, lundberg2017unified, camburu2018esnli, thorne2019generating, rajani2019explain, jain2019attention}.

In this paper, we investigate explainable LM-based models for multi-choice question answering (MC-QA). Specifically, we address the question: \emph{What is the knowledge in the LM used for answering a question?}
Our approach consists of a \emph{generator} and a \emph{classifier} (see Figure~\ref{fig:overview}). The generator takes a question and outputs a \emph{textual hypothesis}, which consists of a few words in natural language. The classifier takes the hypothesis (and possibly the question) to make the final prediction. Unlike end-to-end models where reasoning is internal to the model, the hypothesis is an inspectable intermediate representation that exposes relevant knowledge extracted from the LM. To the best of our knoweldge, we are the first to propose this \emph{intermediate textual layer} for as an explanation for a QA model.

Our setup can be viewed as an instance of \emph{controlled text generation}. However,
generation is not controlled by additional inputs \cite{kikuchi2016controlling, ficler2017controlling, keskar2019ctrl, dathathri2019plug}. Instead, we train from weak supervision -- the generator is trained to produce hypotheses that are useful for the downstream QA application. We compare this approach  to existing methods for generating explanations, where the explanation is provided as a target and the model is trained to generate it explicitly \cite{camburu2018esnli, rajani2019explain}.

Training from weak supervision raises several technical challenges. First, 
because the generator outputs discrete symbols, the loss is non-differentiable with respect to its parameters. We use the Gumbel-Softmax straight-through estimator \cite{jang2016categorical, maddison2016concrete} to overcome this difficulty. Second, the classifier can choose to ignore the generated hypothesis, or it can coordinate with the generator to change the original meaning of words. In such case, the hypothesis will not constitute a useful explanation. To encourage meaningful hypotheses, we (a) train the classifier along with an auxiliary similarity classifier that must use the hypothesis, (b) add loss terms that encourage the hypothesis to correspond to natural language, and (c) feed the classifier with multiple generator outputs.

We evaluate our approach in multiple setups. First, in a synthetic setup, where the model learns to output the hypernym of objects; Second, on \csqa{}, which focuses on commonsense knowledge; and third, on zero-shot transfer to \qasc{}, a multi-choice QA task with an emphasis on scientific knowledge. We find that our approach reaches performance that is comparable to end-to-end models, while providing good hypotheses that are shown to be used by the classifier for prediction.
We analyze the generated hypotheses and demonstrate that they shed light on different reasons for model error in different questions (e.g., missing world knowledge vs. language understanding difficulties), and help detect examples where the prediction of the model does not reflect its true knowledge.
Our approach can be generalized outside of MC-QA to any scenario where we want to control the text generated from a LM using a signal from a downstream application. Our code and data is available at \url{https://github.com/nika2312/qa_explaination}.

%% file: 02_background.tex
\section{Background}
\label{sec:background}



\paragraph{Problem setting} We consider the task of multi-choice question answering, where given a question $q$ and a list of candidate answers $\sA = (a_1, \dots, a_n)$, our goal is to choose the correct answer $a^* \in A$. At training time, we observe question-candidates-answer triples $\{(q_i, \sA_i, a^*_i)\}_{i=1}^N$, from which we train a model.

A standard end-to-end approach for MC-QA is to first obtain a contextualized representation for the question and each answer candidate $a_i$, by passing it through a transformer-based pre-trained LM \cite{vaswani2017attention}:  $h^i = \text{LM}(q, a_i)$. Then, the contextualized representations are summarized to a single vector $g_i$. For example, in BERT \cite{devlin2018bert} the summary vector is $g_i = h^i_\texttt{CLS}$, where $h^i_\texttt{CLS}$ is the representation of the special \texttt{CLS} token. Last, a weight vector $w$ is learned to compute the score $s_i = w^\top g_i$. The final output distribution is $\text{softmax}(s_1, \dots, s_n)$, and the entire model is trained to maximize the log-likelihood of $a^*$.

This simple setup has been successful in multiple MC-QA tasks that require commonsense reasoning \cite{sap2018atomic, zellers2018swag, chen2019codah, talmor2018commonsenseqa, wang2019does}. However, in end-to-end models, all reasoning is internal to the model, and it
is difficult to know what knowledge inside the LM was used for prediction. 




In this work, we develop a model that exposes more directly the knowledge inside the LM that is used, by having it generate this information in the form of language (Figure~\ref{fig:overview}). Our goal is not to improve \emph{performance}, as we assume backpropagation over a differentiable model is an effective method for distilling information from a  pre-trained LM. Instead, we aim for \emph{explainability}, aiming to help both users and practitioners understand 
what the model is doing. Recent work has shown \cite{petroni2019language, jiang2019know, talmor2019olmpics} that it is possible to extract information encoded inside a LM by directly training it to output certain tokens in certain positions. Here, we make a weaker assumption that we do not know what words the model should generate, only that the text should be useful for the downstream application.



%

%% file: 03_model.tex
\section{Model}
\label{sec:model}

At a high-level our model consists of two components, the \emph{generator} and the \emph{classifier}. The generator takes a question $q$ as input, and outputs a hypothesis, which is a sequence of tokens  $c = (c^1, \dots, c^{|c|})$. The classifier takes the question, generated hypothesis, and answer candidates $\sA$, and predicts the answer. 

We assume the generator is based on a large pre-trained autoregressive LM (encoding world knowledge in its parameters). 
The LM provides a distribution over the vocabulary  $p_\text{gen}(x_i \mid x_{1\dots i-1})$, where $x_{1\dots i-1} = (x_1, \dots, x_{i-1})$ is a sequence of observed tokens. Thus, the hypothesis $c$ is generated by decoding the hypothesis left-to-right, concatenating the decoded prefix to the question $p_\text{gen}(c \mid q) = \prod_{i = 1}^{|c|} p_\text{gen}(c_i \mid [q ; c_{1\dots i-1}])$.
In this work, we utilize two well-known LMs, GPT-2 \cite{radford2019language} and XLNet \cite{yang2019xlnet} (see \S\ref{sec:experiments}).

The classifier is very similar to the end-to-end architecture described in \S\ref{sec:background}. It takes the question $q$, the hypothesis $c$, and an answer candidate $a_i$ as input, and outputs a score $s_i$. The final distribution is $\classifier(a \mid q, c) = \text{softmax}(s_1, \dots, s_n)$. We discuss possible forms for the classifier and the effect on performance and explainability in \S\ref{sec:classifier}.


Our framework raises a few technical challenges. First, because we sample discrete symbols from the generator, the log-likelihood of the correct answer $a^*$ is not differentiable with respect to the parameters of the generator $p_\text{gen}$. While this indeed makes optimization challenging, we overcome this by using the straight-through Gumbel-softmax estimator \cite{jang2016categorical, maddison2016concrete}, which is a standard approach in this setup (\S\ref{sec:optimization}). Second, if the classifier is strong enough, for example, another instance of a large pre-trained LM, it can easily answer the question directly, completely ignoring the hypothesis $c$ generated by the generator. To overcome this, we train a much simpler similarity  classifier (possibly jointly with a more complex classifier), which provides incentive for the generator to produce useful hypothesis (\S\ref{sec:classifier}). Last, how do we make sure that the hypothesis is a useful explanation for humans? We experiment with both loss functions and decoding mechanisms that improve the quality of the hypothesis (\S\ref{sec:explainability}).

\subsection{Training and Optimization}
\label{sec:optimization}
Our goal is to obtain a good MC-QA model, and thus we maximize the expected log-likelihood of the correct answer:
\begin{equation} \label{eq:objective}
\mathbb{E}_{\hat{c} \sim p_\text{gen}(c \mid q)} [ \log p_\text{QA}(a^* \mid q, \hat{c}) ].
\end{equation}
Computing the expectation exactly is intractable, and so we approximate it by sampling $\hat{c}$. 

Training the classifier is trivial, since the loss is differentiable with respect to its parameters. However, the loss is not differentiable with respect to the parameters of $p_\text{gen}$. To overcome this difficulty, we use the straight-through (ST) Gumbel-softmax (GS) estimator, which has been shown to produce better gradient estimates compared to REINFORCE \cite{williams1992simple}. The Gumbel-softmax trick provides a continuous relaxation for the categorical distribution over the vocabulary, making the model fully differentiable. However, it results in a mismatch between training and test time, as at test time we want discrete hypotheses. The straight-through estimator solves that by using \texttt{argmax} in the forward pass and \texttt{softmax} in the backward pass. For details on the GS-ST estimator, please see \newcite{bengio2013estimating,jang2016categorical, maddison2016concrete, yin2019understanding}.


We note that because the loss is backpropagated through the inputs of the classifier to the outputs of the generator, the vocabularies of the generator and classifier must be tied. In practice, we use XLNet (or GPT-2) for both generation and classification.


\subsection{Classifier Expressivity}
\label{sec:classifier}

We would like to have hypotheses that reflect knowledge in the LM generator that is useful for the MC-QA task. However, if the classifier has the same knowledge as the generator (for example, if they are initialized with the same parameters), it has no incentive to use the hypothesis at all, as it can extract the same knowledge from its own weights. Moreover, even if the classifier is weaker than the generator, once it has enough capacity, it can coordinate with the generator in arbitrary ways, and make the tokens decoded by the generator lose their original meaning. Indeed, in \S\ref{sec:experiments} we show that in such cases hypotheses are meaningless. 

In this work, we consider a simple \emph{similarity classifier} $\weakclassifier$ whose only parameters are word embeddings (excluding the generator parameters). Specifically, given the sequence of word embeddings for the context $E_c = (e_c^1, \dots, e_c^{|c|})$ and the sequence of word embeddings for an answer candidate $E_{a_i} = (e_a^1, \dots, e_a^{|a_i|})$, we define the score for the candidate answer as:
$$
s_i = \frac{1}{|c||a_i|}\sum_{j=1}^{|c|}\sum_{k=1}^{|a_i|} {{e_c}^j}^\top e_a^k = \text{avg}(E_c^\top E_{a_i}).
$$
This model does not utilize the question at all and hence must use the hypothesis $c$ to answer the question. This pushes the generator towards generating interpretable hypotheses that are more similar to the true answer compared to the distractors.

The similarity classifier $\weakclassifier$ encourages the generator to produce meaningful hypotheses, but this can come at a great cost to performance. Thus, we propose to train $\weakclassifier$ jointly with a more expressive LM-based classifier $\strongclassifier$. Specifically, this classifier is another large pre-trained LM almost identical to the end-to-end model described in \S\ref{sec:background}. The only differences are that (a) it takes as input the concatenation of the question, answer candidate, \emph{and the generated hypothesis}, (b) the summary vector $g_i$ encoding the input is the last hidden state of the contextualized representation. The classifiers share word embeddings and receive the same generator's output. The modified objective is hence:
\begin{equation}
\label{eq:multitask}
\mathbb{E}_{\hat{c} \sim p_\text{gen}(c \mid q)} [ \log (\weakclassifier(a^* \mid q, \hat{c}) \strongclassifier(a^* \mid q, \hat{c}) ) ].    
\end{equation}
We empirically show in \S\ref{sec:experiments} through ablation tests that this \emph{LM-based classifier} indeed uses the hypotheses $c$ generated by the generator, even when it is given the question as input.

Last, we note that $\weakclassifier$ pushes the generator to produce hypotheses that are similar to the answer. In other setups, we might want the generator to produce hypotheses that complement a different source of information. For example, if the question is accompanied by a paragraph that contains relevant information. In this case, the classifier can be a reading comprehension (RC) model whose parameters are untrained, and the generator will be pushed to produce hypotheses that allow the RC model to answer the question.

\subsection{Explainability}
\label{sec:explainability}




The similarity classifier encourages the generator to generate meaningful hypotheses. However, there is no guarantee that the meaning of words does not change during training. Moreover, the similarity classifier objective might lead to repetitions in the generated text. We now describe variants aimed at improving interpretability.


\paragraph{KLD loss}
We add a KLD loss regularization term to the generator objective \cite{jaques2016sequence}: given the original pre-trained LM $p_\text{NL}$, we minimize at each decoding step the KL divergence $D(p_\text{gen} \vert\vert p_\text{NL})$ between the trained generator distribution and the original pre-trained LM distribution.
This prevents drift of the generator distribution.

\paragraph{Repetition penalty}
We adopt the unlikelihood objective from \newcite{welleck2019neural} to discourage repetitions in the generated text. For each decoded token $c_i$, we add to the loss function the term $\sum_{w \in \sW^t}{\log(1-p_{\text{gen}}(w \mid [q ; c_{1...i-1}))}$, where $\sW^t$ is a set of ``negative" tokens we want to penalize, that is, those that already appeared in $c_{1...i-1}$.


\paragraph{Top-$K$ sampling}
Sampling a single hypothesis $\hat{c}$ provides a narrow channel for the generator to pass information about its distribution to the classifier. To provide more information, we modify the ST estimator forward pass, and output the top-$K$ tokens before passing the hypotheses to the classifier. This produces a set of similar words that are guaranteed to be distinct from one another. To avoid the computational burden of running the classifier over a beam of hypotheses, we perform just one step of decoding, and concatenate the top-$K$ tokens with highest probability (thus, $|c|=K$).



\subsection{Supervised Generator}
Several recent works learned to generate
answers or explanations \cite{mccann2018natural, camburu2018esnli, rajani2019explain, huang2019cosmos} using the standard supervised sequence-to-sequence objective \cite{sutskever2014sequence}. Specifically, they encode a source sequence and maximize the log-likelihood of the target explanation (or answer) token-by-token using cross-entropy loss over the vocabulary. We adapt this approach to our setup, and encode the question $q$ as usual, but use the correct answer $a^*$ as the target. Here, the model obtains supervision in every decoding step, unlike weak supervision where the loss is obtained once for the entire hypothesis. We refer to this model variant in \S\ref{sec:experiments} as \supgen{}.\footnote{Training \supgen{} jointly with the LM-based classifier resulted in poor performance, due to optimization issues.}



%% file: 05_experiments.tex
\section{Experiments}
\label{sec:experiments}

We evaluate on the synthetic task of hypernym extraction, on \csqa{}, and on transfer to \qasc{}.

\subsection{Sanity Check: Hypernym Extraction}
\label{sec:sanity}

We investigate whether our model can extract hypernyms, which have been shown to be encoded well in LMs \cite{Richardson2019WhatDM}.

We automatically create a dataset of 7,625 questions of the form \nl{What is a \texttt{[hyponym]}?}, where \texttt{[hyponym]} is a slot filled by words, such as \nl{dog}, and the possible answers are six disjoint hypernym categories: (a) \nl{plant}, (b) \nl{bird}, (c) \nl{fish}, (d) \nl{mammal},  (e) \nl{reptile}, and (f) \nl{bacteria}. Hyponym-hypernym pairs are harvested from ConceptNet \cite{speer2016conceptnet}. We create a development set by randomly sampling 20\% of the data, and use the rest for training.

We use GPT-2 as a generator, generating a single word ($|c|=1$), and optimize the
similarity classifier only: $\mathbb{E}[ \log p_\text{QA}^\text{sim}(a^* \mid q, \hat{c})$. Our goal is to compare this with a standard end-to-end (\S\ref{sec:background}) model, termed \etoebaseline{}.
We observe that the accuracy of our model is 84.0, close to the accuracy of \etoebaseline{}, at 86.5. This shows that the model is able to convert the knowledge in its parameters into language tokens. More results including the effect of the components of the GS-ST estimator are in Appendix~\ref{sec:hypernym}.


\begin{figure}[t]
 \includegraphics[width=1\columnwidth]{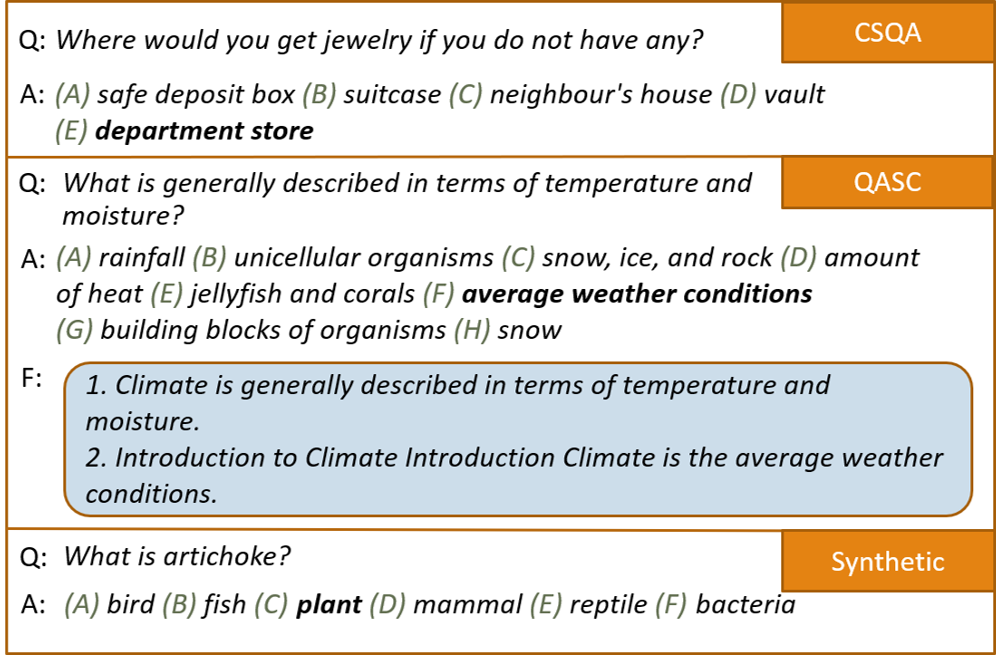}
 \caption{Examples from the different datasets we use.
 }~\label{fig:datasets}
\end{figure}

\subsection{MC-QA Experiments}
\label{sec:mcqa}

We evaluate our models on \csqa{} \cite{talmor2018commonsenseqa}, a multi-choice commonsense reasoning QA benchmark consisting of 12,102 questions with 5-choice answers, and on zero-shot transfer to \qasc{} \cite{khot2019qasc}, a dataset containing science questions that require external knowledge, consisting of 9,980 examples with 8-choice answers (Figure~\ref{fig:datasets}). 
\qasc{} contains a corpus of scientific facts for information retrieval (IR) purposes, but we do not use it in our setup.

All models are trained for 20 epochs, using BertAdam optimizer with a learning rate of $2e^{-5}$, batch size of 8, and dropout of 0.1. Training was done using AllenNLP \cite{gardner2018allennlp}, on top of models from Hugging Face \cite{wolf2019huggingfaces}.

\subsubsection{Similarity classifier experiments}
\label{sec:sim-classifier}
We start by examining the performance and explainability of models trained using a XLNet-based similarity classifier only (Eq.~\ref{eq:objective}), and examine the effect of the variants from \S\ref{sec:model}.

\paragraph{Baselines} The aforementioned \etoebaseline{} baseline (\S\ref{sec:sanity}) obtains an accuracy of 71.0 on the development set of \textsc{CSQA}. However, it has two advantages compared to our approach that we wish to disentangle: (a) it is fully differentiable (this is what we wish to isolate), but also (b) its architecture is different: it is given both the question and answer candidate as inputs to a transformer. Thus, it can model arbitrary interactions between them at the token level. This is in contrast to our model, where the generator only observes the question, and generates a hypothesis that is used to deterministically select the answer. Thus, we introduce the \nointeraction{} baseline, which is fully differentiable, but has limited interaction between the question and answer candidates. 

In \nointeraction{}, we feed only the question $q$ as input to the pre-trained LM, and represent it with the last contextualized representation $g_\text{final}$. Then, the score for an answer candidate is computed just like in the similarity classifier: $s_i = \text{avg}(g_\text{final}^\top E_{a_i})$. This results in an end-to-end model without a hypothesis, where the question and answer candidates have limited interaction.
\nointeraction{} obtains 63.7 accuracy on the development set, showing that with this limited interaction performance drops by 7-8 points. We view this performance as an upper bound on the accuracy of the similarity classifier.


\begin{table}[t]
\centering
\footnotesize
\resizebox{0.92\columnwidth}{!}{
\begin{tabular}{lcc} \hline\hline
Model               & Accuracy  & \% repetitions \\ \hline
\textsc{$|c|=1$}          & 53.3 &  -\\
\textsc{$|c|=3$}          & 54.0   & 63\\
\textsc{$|c|=3$+KLD}          & 51.3   & 42\\
\textsc{$|c|=3$+KLD+REP}          & 49.0  & 19\\
\textsc{$|c|=5$}          & 52.8    & 78\\
\textsc{$|c|=5$+KLD}          & 52.2   & 68\\
\textsc{$|c|=5$+KLD+REP}          & 50.7  & 14\\
\textsc{Top-$K=3$ ST}          & \textbf{58.0}  & - \\
\textsc{Top-$K=5$ ST}          & 56.2    & -\\
\hline
\supgen{} $|c|=3$ & 50.8 & 0.9 \\
\hline
Comparable \etoebaseline{} & \textbf{63.7} & - \\ \hline\hline
\end{tabular}}
\caption{Similarity classifier accuracy on the development set. \% repetitions corresponds to the number of repeated words in a hypothesis across all examples.}
\label{tab:weakresults}
\end{table}

\paragraph{Results} Table~\ref{tab:weakresults}  shows the performance of different generator models, where we compare different hypothesis lengths ($|c|$), use both the \textsc{KLD} and repetition (\textsc{REP}) loss functions, and the ST estimator with \textsc{Top-$K$} sampling. Table~\ref{tab:gen_output} presents examples of hypotheses generated by the different generators that highlights the differences in explainability between them.

Our best model is \textsc{Top-$K=3$ ST} that reaches 58.0 accuracy, and by design outputs three different words in one decoding step. This is 5.7 points lower than  \nointeraction{}, and we attribute the lower performance to the narrow channel between the generator and classifier (words instead of vectors), and to the difficult optimization.

Models with multiple decoding steps tend to produce repeated words, due to the nature of the similarity classifier, which encourages generating words that are similar to the correct answer. Adding the KLD and especially the repetition loss reduces repetitions dramatically, but also decreases accuracy. Table~\ref{tab:gen_output} shows that the generated text is reasonable, even when different from the gold answer (we evaluate and analyze the quality of hypotheses more explicitly below).
Surprisingly, KLD loss did not yield more coherent phrases, perhaps due to the known tendency of LMs to generate repetitions \cite{holtzman2019curious, welleck2019neural}. 

We evaluated the supervised generator, \supgen{}, outputting exactly 3 tokens and then applying the similarity classifier.
\supgen{} produces text that is more natural than our models, as evidenced by the examples in Table~\ref{tab:supgen_output}. However, because it is not optimized for QA, accuracy drops by 7.2 points.





\subsubsection{LM-based classifier experiments}

\begin{table}[t]
\centering
\footnotesize
\begin{tabular}{lccc|c} \hline\hline
Generator           & \textsc{Q+C} & \textsc{Q} & \textsc{$\Delta (\%)$} & \qasc{}  \\ \hline
\weaksup{}    &   70.0 & 70.0 & 0.0 & 37.0 \\
\hline
\textsc{$|c|=1$}          & 69.0   & 62.0 & -10.1 & 35.1\\
\textsc{$|c|=3$}          & 69.0   & 63.5 & -8.0 & 35.7\\
+\textsc{KLD}          & 70.0   & 63.5 & -9.3 & 38.4\\
+\textsc{KLD+REP}          & 70.0   & 65.0 & -7.1 & 36.4\\
\textsc{Top-$K=3$ ST}          & 70.0   & 60.4 & \textbf{-13.7} & 35.7\\
\textsc{Top-$K=5$ ST}          & \textbf{70.9}    & 60.7 & -13.1 & \textbf{39.2}\\
\hline
\supgen $|c|=3$ & 66.7 & 56.6& -15.4 & - \\
\supgen $|c|=30$ & 66.7 & 48.4& \textbf{-27.4} & - \\
\hline
\etoebaseline{}             & -   & 71.0  & - & 38.9\\ \hline\hline
\end{tabular}

\caption{Accuracy on the development set of the LM-based classifier, trained jointly with the similarity classifier; \textsc{Q+C}: The classifier takes the question and hypothesis as input at training and test time; \textsc{Q}: The classifier takes the question and hypothesis as input at training time, but only the question at test time; $\Delta$: performance drop (in $\%)$ between \textsc{Q} and \textsc{Q+C};
\supgen{} was trained independently from the classifier to generate the answer $a^*$.
\qasc{}: Accuracy when trained on \csqa without further fine-tuning.} 
\label{tab:strongresults}
\end{table}

We now examine the performance and explainability of joint training of the LM-based classifier and the similarity classifier (Eq.~\ref{eq:multitask}). To show the importance of joint training, we evaluate \weaksup{}, where the model is trained without the similarity classifier. When training jointly, we start with the similarity classifier only, as a warm-up for the generator, and then add the LM-based classifier.

Table~\ref{tab:strongresults} shows the results. The \textsc{Q+C} column shows results when the classifier is trained and tested using the question and hypothesis. The \textsc{Q} column shows results when training with the question and hypothesis as usual, but at test time only the question is passed, and the \emph{hypothesis input is zeroed out}. This indicates whether the LM-based classifier uses the hypothesis or ignores it. 

We observe that \weaksup{} gets high accuracy, comparable to \etoebaseline{}. However, because the classifier is strong, it can ignore the hypothesis, and generated hypotheses are meaningless. This is also evident by the fact that zeroing out the hypothesis does not change performance.

All weakly-supervised generators reach roughly the same accuracy, and their performance dramatically drops in lieu of the hypothesis, showing that the LM-based classifier uses it. \textsc{Top-$K=5$ ST} performs best --  almost the same as \etoebaseline{}. However, models differ in terms of
how much they rely on the hypothesis (columns \textsc{Q} and $\Delta$). Performance drops by more than 13\% for both \textsc{Top-$K=3$ ST} and \textsc{Top-$K=5$ ST}, indicating that the LM-based classifier strongly relies on the generated hypotheses. Table~\ref{tab:gen_output} shows some generated hypotheses, which are similar to those obtained when training with a similarity classifier only (\S\ref{sec:sim-classifier}).



\begin{table}[t]
\centering
\footnotesize
\begin{tabular}{lc} \hline\hline
Model               & Score  \\ \hline
\textsc{$|c|=3$+KLD+REP}          & 0.72  \\
\textsc{Top-$K=5$ ST}          & \textbf{0.74}   \\
\supgen{} $|c|=3$ & 0.60  \\
\supgen{} $|c|=30$ & 0.55 \\ \hline\hline
\end{tabular}
\caption{Human-evaluation results for how reasonable hypotheses are (\csqa{} development set). 
Each rater determined whether a hypothesis is reasonable (1 point), somewhat reasonable (0.5 point) or not reasonable (0 points). The score is the average rating across raters and examples.}
\label{tab:human_eval}
\end{table}


\begin{sidewaystable}
\tiny
{
\begin{tabular}{p{5cm}p{1cm}p{1cm}p{2.5cm}p{2.5cm}p{2.5cm}p{3.5cm}p{3.5cm}} \hline\hline
Question & gold & $|c|=1$ & $|c|=3$ &  \textsc{$|c|=3$+KLD+REP} & \textsc{$|c|=3$+KLD+REP E2E} & \textsc{Top-$K=5$} & \textsc{Top-$K=5$ E2E} \\ \hline
\nl{what would use a musical instrument?} & orchestra & entertainment & play play play & music Music Music & concert movies opera & band orchestra music concert sing & band concert orchestra music show \\
\nl{where would you find a ticket booth and see a concert?} & venue & theatre & theater theatre theatre & concert Theater concert & theater theaters theater & theatre theater auditorium arena stadium & theater theatre auditorium stadium venue \\
\nl{what do people do when they don't understand something?} & ask questions & question & questions question questions & criticize odi question & argue Questions questions & questions question study ask think & questions confusion confused puzzled confuse \\
\nl{where are required to carry books all day?} & university & school & classroom classroom classroom & office offices office & classrooms backpack backpack & classroom library classrooms school libraries & library school libraries office librarian \\
\nl{they had a theory of what they could do in the big game, so over and over they would what?} & practice & practice & attack attack attack & practice Practice practice & practice Strategies strategy & practice play try practiced execute & strategy practice think strategies drill \\
\nl{where might an unused chess set be stored?} & cupboard & cupboard & cupboard cupboard cupboard & cabinet cupboard cabinet & cupboard closet closet & cupboard closet garage drawer cabinet & cupboard drawer cabinet closet basement \\
\nl{a human wants to submerge himself in water, what should he use?} & whirlpool bath & bath & bathtub bathtub bathtub & bath Bath pool & bath submerged sponge & bath bathtub tub pool bathroom & bath bathtub tub sink shower \\
\nl{where can you put a picture frame when it's not hung vertically?} & table & shelf & shelf shelf shelf & shelf shelves shelf & shelf shelves shelf & shelf table floor drawer shelves & shelf floor table ceiling shelves \\
\nl{because john was first violin, he had to bring something important to work ever day. what did he need to bring to work?} & violin case & violin & instruments instruments instruments & briefcase briefcase briefcase & violin Instruments violin & instruments instrument violin briefcase backpack & violin briefcase backpack instrument laptop \\
\nl{what do people aim to do at work?} & complete job & achieve & accomplish work work & achieve achievement advancement & improve work work & achieve produce results accomplish do & progress accomplish improve customers improvement \\
 \hline\hline
\end{tabular}
}
\caption{Hypotheses generated when training the similarity classifier. $|c|$ indicates the hypothesis length; \textsc{KLD} and \textsc{REP} correspond to KLD loss and repetition loss; \textsc{E2E} indicates joint training with the LM-based classifier.}
\label{tab:gen_output}
\end{sidewaystable}

Tables~\ref{tab:fix} and~\ref{tab:break} (Appendix~\ref{sec:examples}) present examples where zeroing out the hypothesis input (column \textsc{Q}) creates or fixes an error for \textsc{Top-$K=3$ ST}.
We observe how the hypothesis sways the prediction of the model, and that often even when an error is caused, the hypothesis is reasonable.


We evaluate \supgen{}, which was trained independently as a sequence-to-sequence model and is not optimized for QA. We experiment with decoding a hypothesis of length $|c|=3$ and also of maximal length $|c|=30$. Results show that indeed performance is 4 points lower than our best model, but the output text is more natural. Moreover, the LM-based classifier relies on it for its prediction, especially when we decode very long hypotheses $|c|=30$. To summarize, there is a trade-off between the two types of supervision, where training from the QA signal leads to higher accuracy, but text that is ``list-like" and not natural (Table~\ref{tab:gen_output}), while training to generate the answer yields lower performance, but more natural text (Table~\ref{tab:supgen_output}).


\paragraph{Human evaluation}

To better understand the quality of the generated text, we perform a human evaluation. We randomly sample 50 examples from the development set, and generate hypotheses from four models: (a) $|c|=3$+\textsc{KLD}+\textsc{REP}, (b) \textsc{Top-$K=5$ ST}, (c) \supgen{} $|c|=3$ and (d) \supgen{} $|c|=30$. We randomly shuffle question-hypotheses pairs from all models, show them to six graduate students, and ask them to rate whether a hypothesis is reasonable, somewhat reasonable, or not reasonable. Each hypothesis is rated by 3 students, and the score for a model is the average across raters and examples.

Table~\ref{tab:human_eval} shows the results of this experiment. Top-$K=5$ ST achieved the highest score of 0.74. While \supgen{} models produce more natural texts, they are judged to be less reasonable in the context of the question.


\paragraph{Zero-shot transfer to QASC}
We examine whether our model can generalize to the \qasc{} dataset, where general knowledge is needed.
An \etoebaseline{} model trained on \csqa{} obtains 38.9 accuracy (Table~\ref{tab:strongresults}), showing that a model trained on \csqa{} transfers reasonably well to \qasc{} without fine-tuning. 
Our hypothesis-generating models also generalizes without fine-tuning to \qasc{}, where \textsc{Top-$K=5$ ST} reaches the highest accuracy, 39.2, while also providing the hypotheses as explanation.
Table~\ref{tab:qasc_output} shows examples for generated hypotheses for \qasc{} questions.

\paragraph{\csqa{} Test set results}
We evaluated \textsc{Top-$K=5$ ST} on the test set of \csqa{} and obtained 63.5 accuracy. As a point of reference, the leaderboard of \csqa{} reports one model that uses XLNet-large, which obtains 66.9 accuracy, but also uses external  documents retrieved from Wikipedia with IR.\footnote{DREAM entry on \url{https://www.tau-nlp.org/csqa-leaderboard}.}


\begin{table*}[t]
\centering
\resizebox{1.0\textwidth}{!}{
\begin{tabular}{lll}
\hline \hline
Question & Hypothesis  & Answer candidates\\ \hline
\nl{what mall store sells jeans for a decent price?} &
store mall stores retail shop &
apartment, \textbf{GAP}, bedroom, \underline{thrift store}, clothing store   \\
\nl{the gimmicky low brow TV show was about animals when they what?} &
animals humans human aliens live &
sick, males, \underline{mammals}, bite, \textbf{attack}  \\ \hline
\nl{where would you use a folding chair but not store one?}  
& cupboard cabinet closet drawer room  & \underline{closet}, \textbf{beach}, city hall, school, garage \\
\nl{where would you get some maps that you own?} & store stores bookstore shop Store & important when traveling, library, \textbf{cabinet}, electrical circuit, \underline{bookstore}   \\ \hline
\nl{athletes soak in hot tubs to relieve what after playing baseball?} &
sweat pain stress headache exhaustion &
strikes, \underline{pain}, errors, fame, \textbf{sore muscles} \\
\nl{where would you find a monkey in the wild?} &
wild woods bush range forest &
\textbf{Thailand}, captivity, barrel, research laboratory, \underline{zoo}  \\ \hline \hline
\end{tabular}}
\caption{Top: examples for bad hypotheses due to \emph{missing knowledge}. Middle: 
Examples for bad hypotheses due to semantic errors. Bottom:
Examples for reasonable hypotheses that do not fit the specific distractors.
\textbf{Bold}: gold answer; \underline{underline}: predicted answer.}
\label{tab:analysis}
\end{table*}

\subsection{Explainability Analysis}
\label{sec:analysis}

We analyze how the textual hypotheses provide insights onto the abilities of the LM beyond what is possible with an end-to-end architecture. We analyze our joint model \textsc{Top-$K=5$ ST}, and present examples in Table~\ref{tab:analysis}.

First, we look at cases where both the similarity and LM-based classifiers answered correctly (49\% of the cases). We manually annotated whether the hypotheses provide a reasonable answer for a random sample of 100 hypotheses from the development set, assigning 1 point to a reasonable hypothesis, 0.5 point to a somewhat reasonable hypothesis, and 0 points to an unreasonable hypothesis. The score was 0.94, indicating that when both classifiers are correct, the hypotheses are reasonable, and we can have confidence that the model does not ``cheat".
In the few cases where the hypothesis is unreasonable, it reveals a shortcoming in the model's knowledge. For example, for the question \nl{The hostess was good at her job, she always had a smile when she would what?}, the model outputs the hypothesis \nl{dinner eat serve food meals}, indicating that it does not distinguish \emph{a hostess} from \emph{a waitress}. However, the distractors are weak, and the model correctly chooses \nl{welcome guests}.

Next, we look at cases where both classifiers were wrong (23\% of the cases). Annotating 51 examples to estimate whether hypotheses are reasonable, we obtain a very low score of 0.21. By examining the hypotheses, we can understand the main reasons of error (examples in Table~\ref{tab:analysis}): (a) \emph{Missing knowledge}: the question requires very specific knowledge that the model does not generate, suggesting that the knowledge is missing. (b) \emph{Semantic errors}: the model ignores parts of the question and is misled by surface clues (ignoring negation in the first example, and a restrictive relative clause in the second).  These two reasons cover 54\% of the unreasonable hypotheses produced, showing how the hypotheses provide important information for ``debugging" the model.

We also analyze cases where the LM-based classifier was correct but the similarity classifier was wrong (22\% of the cases). Annotating 50 such examples produces a score of 0.55, showing that many of the hypotheses were actually reasonable. We observe that in two-thirds of these cases, the error is related to the inability of the generator to consider the distractors themselves, and while the hypothesis was reasonable, it was more similar to a distractor than to the gold answer (Table~\ref{tab:analysis}).
Finally, transferring to \qasc{} (Table ~\ref{tab:qasc_output}) shows that the model performs well on general knowledge questions, which are somewhat similar to those in \csqa{}, but fails on more scientific questions that require encyclopedic knowledge.

\paragraph{Summary}
We observe that joint training leads to a model that has comparable performance to an end-to-end model, but also provides a window to the information inside the LM. Specifically, using the hypotheses we can detect cases where the model was right, but lacks the necessary knowledge, cases where the model was wrong, but provides reasonable answers, and analyze different types of failures related to knowledge and language.

%% file: 06_related.tex
\section{Discussion and Related Work}


\paragraph{Explanation datasets} 
There has been substantial effort recently to collect human-generated explanations and train models to generate or select them.
\newcite{rajani2019explain} presented the CoS-E dataset, where human-generated explanations to commonsense questions are used to train a LM. \newcite{wang2019does} created a dataset for evaluating whether a model can choose the right explanation for its decision. \newcite{huang2019cosmos} crowd-sourced a multi-choice reading comprehension dataset, where answer candidates are human-generated explanations, and showed that a trained generative model can produce semantically consistent explanations. \newcite{sap2019socialiqa} created a dataset for explaining social situations.
All these approaches rely on human-generated explanations that are expensive to collect. Moreover, it is unclear whether generated explanations are actually used for predicting the correct answer.

\paragraph{Explainability} There has been ample research recently on precisely defining what are explanations and what are their different facets (see \newcite{lipton2016interpretability} and \newcite{wiegreffe2019attention} among others). Specifically, an important question is whether the explanation \emph{causes} the prediction, that is, whether the model actually uses the explanation to reach its prediction. From this perspective, our similarity classifier provides hypotheses that strongly influence the prediction, as the question is not even passed to the classifier. Our LM-based classifier can potentially choose to ignore the hypothesis, but we show experimentally that ablating the hypothesis results in a performance drop. Thus, it is more similar to attention as an explanation \cite{serrano2019attention,jain2019attention,pruthi2019learning}, where the attention structure does not necessarily reveal what tokens are used for prediction. Our supervised model is  an even weaker form of explanation since it is trained independently from QA.

\paragraph{Multi-choice QA}
MC-QA has been a popular format for QA \cite{lai2017race, zellers2018swag, clark2018think} mostly because it simplifies \emph{evaluation} of free text answers. However, users in the real world ask questions in order to get an answer, and not to test the knowledge of a model or student \cite{chen2019evaluating}. Thus, it is difficult to know whether models that succeed in MC-QA can actually answer the question, as they may take advantage of weaknesses in the way distractors were constructed. Our similarity classifier model can be viewed as an \emph{abstractive QA} model that answers the question directly, and is evaluated with a multiple-choice format. In future work, our similarity classifier can be replaced by more complex untrained models that choose the right answer given the generated hypothesis.




%% file: 07_conclusion.tex
\section{Conclusion}
In this work with propose a LM-based model for MC-QA, which generates natural language text that can be used to understand the the knowledge extracted from the LM,  as an intermediate step. The performance of our model is on par with end-to-end models, while providing an inspectable layer for practitioners and users. Our approach is supervised from downstream application signal only, and thus can be generalized to any scenario where we would like to train a LM to generate text that is useful for a downstream application.

%% file: 08_rotated_tables.tex
\clearpage
\begin{sidewaystable}[H]
\tiny
\centering
{
\begin{tabular}{p{8cm}p{2cm}p{4cm}p{8cm}} \hline\hline
Question & gold &  \supgen{} $|c|=3$ & \supgen{} $|c|=30$ \\ \hline
\nl{what would use a musical instrument?} & orchestra & band practice room & play music hall , and orchestra music  , organ o ob o .\\
\nl{where would you find a ticket booth and see a concert?} & venue  & auditorium building outside & arena building or , and theatre building or auditorium building building on campus of college or school building campus\\
\nl{what do people do when they don't understand something?} & ask questions  & ask questions & obsess , and obsess over  .\\
\nl{where are required to carry books all day?} & university &  school backpacks & classroom building or , and school backpack s are required to carry textbooks all day long .\\
\nl{they had a theory of what they could do in the big game, so over and over they would what?} & practice  & repeat attacks . & practice hard and , and and\\
\nl{where might an unused chess set be stored?} & cupboard & cabinet drawers & cabinets of , and cabinets of  quarrels of wood  , music room storage cabinets\\
\nl{a human wants to submerge himself in water, what should he use?} & whirlpool bath & beach ball sack & swimming pool of , and get wet in water fountains and waterfalls of\\
\nl{where can you put a picture frame when it's not hung vertically?} & table  & shelf  rod & shelf unit drawer, and wall shelf (cabinet drawer).\\
\nl{because john was first violin, he had to bring something important to work ever day. what did he need to bring to work?} & violin case  & violin case, & orchestration apparatus, and\\
\nl{what do people aim to do at work?} & complete job & accomplish goals and & increase income and, and  responsibilities and respect ability in jobs with respect ability standards and pay well enough to keep\\
 \hline\hline
\end{tabular}
}
\caption{Hypotheses generated by \supgen{} $|c|=3$ and \supgen{} $|c|=30$ models.}
\label{tab:supgen_output}
\end{sidewaystable}


\begin{sidewaystable}[H]
\tiny
\centering
{
\begin{tabular}{p{5cm}p{2cm}p{3cm}p{4cm}p{2cm}p{5cm}} \hline\hline
Question & gold & \textsc{$|c|=3$+KLD+REP E2E} & \textsc{Top-$K=5$ E2E} &  \supgen{} $|c|=3$ & \supgen{} $|c|=30$ \\ \hline
\nl{what is described in terms of temperature and water in the air?} & climate & balance temperatures temperature & humidity climate atmosphere air temperature & moisture linger & atmosphere of earth , and planet earth .\\
\nl{what are busses used for?} & Transporting humans & transportation Transport transportation & transportation travel transport traveling Transportation & transportation purpose transportation & transporting people and, and cargo (stuff) between cities and countryside (land masses) .\\
\nl{what do spiders catch?} & insects & insects insect insects & insects grass web leaves insect & webs of & webs of , and spider webs of fish webs web china city honey combs spider webs web michigan\\
\nl{kidney failure may be treated with a way of cleaning what?} & blood & kidney gastrointestinal blood & kidney urine blood colon intestine & blood vessels of & liver cells out, and out of blood vessels. \\
\nl{what is used for navigation?} & a compass & compass Sai stars & GPS satellites radar radio map & navigation & radio transmission system, and satellites, submarine navigation, balloon air navigation, space shuttle flight \\ \hline
\nl{what is a simple mode of transportation?} & With feedback loops & automobile buses bus & taxi car bus motorcycle bicycle & car , bus & bicycles and , and motor cars and airplanes and vehicles of any kind .\\
\nl{what are aquatic?} & anemones & Life organisms underwater & animals creatures organisms animal mammals & life forms like & fishes and, and  aquatic life  ephemeral  eddies of water currents moving fast out of depth to surface water \\
\nl{what do choanocytes have to trap the particles?} & Protein & lung antibodies cell & membrane uterus muscles tubes arteries & cell wall lining & muscles of , and muscles of rat tail paws , a cat'\\
\nl{what traps particles?} & flagellum or tiny hairs & vacuum ! $\langle$/s$\rangle$ & vacuum magnet particle cage trap & container of liquid & solid material barrier , and solid wall of solid wall construction or structure , concrete block wall , wall binder , wall\\
\nl{what is the process by which neurons are created?} & melanin content & differentiation surgical reproduction & growth creation reproduction evolution birth & growth out of & reproduction of life , and humanity being created upon planet earth is fundamental to nature and logical progression of universe as\\
 \hline\hline
\end{tabular}
}
\caption{Hypotheses generated by models trained on \csqa{} and evaluated on \qasc{}. Upper part: examples where the models contained the required knowledge to answer. Lower part: examples where the models lacked the required knowledge.}
\label{tab:qasc_output}
\end{sidewaystable}

%% file: 09_appendix.tex
\clearpage
\section{Examples of the Impact of the Hypothesis on Prediction}
\label{sec:examples}

\vspace{1\baselineskip}

\begin{table}[H]
\centering
\tiny
{
\begin{tabular}{p{6cm}p{2.5cm}p{1.5cm}p{2.5cm}p{2.5cm}}
\hline \hline
Question & Hypothesis & Gold answer     & Predicted with hypothesis & Predicted without hypothesis \\ \hline
\nl{the dogs were protecting their own when they decided to what the bad man?}
& attack chase go & attack & attack & defend \\
\nl{what are you using if there are speakers strapped on your ears?} & radio headphones music & headphones  & headphones & conference \\
\nl{what does everyone have in relation to other people?} &
relationship relationships feelings &
feelings &
feelings &
unique personality \\
\nl{if the president wanted to ban snakes, where would he issue such a decree?} &
congress law Congress &
white house  &
white house &
new mexico \\
\nl{what is the sun ultimately responsible for?}  &
life world existence &
life on earth &
life on earth &
heat \\
\nl{where can you store you spare linens near your socks?} &
closet bedroom drawer &
dresser drawers &
dresser drawers &
home \\
\nl{what would be necessary for getting in shape?} &
exercise fitness exercises &
exercise &
exercise &
good health \\ \hline \hline
\end{tabular}}
\caption{Examples where the hypothesis from \textsc{Top-$K=3$ ST} is useful, and zeroing it out causes an error.}
\label{tab:fix}
\end{table}

\begin{table}[H]
\centering
\tiny
{
\begin{tabular}{p{6cm}p{2.5cm}p{1.5cm}p{2.5cm}p{2.5cm}}
\hline \hline
Question & Hypothesis & Gold answer     & Predicted with hypothesis & Predicted without hypothesis \\ \hline
\nl{what can eating lunch cause that is painful?}
& headache headaches pain & heartburn & headache & heartburn \\
\nl{what happens to a dog before someone puts up posters of them?} & tame trained clean & get lost  & trained & get lost \\
\nl{john was an aristocratic fox hunter. where might he live?} &
England France London &
new hampshire &
england &
new hampshire \\
\nl{where can children play with animals?} &
park zoo farm &
fairgrounds  &
zoos &
fairgrounds \\ \hline \hline
\end{tabular}}
\caption{Examples where the hypothesis from \textsc{Top-$K=3$ ST} hurts the model, and zeroing it out fixes an error.}
\label{tab:break}
\end{table}

\section{Hypernym Extraction}
\label{sec:hypernym}

We report results on the synthetic hypernym extraction task with and without the Gumbel-softmax trick and the ST estimator. Table ~\ref{tab:toy_results} shows the results. We observe that the ST estimator is crucial even on such a simple task, which aligns with prior observations \cite{havrylov2017emergence} that ST helps overcome the discrepancy between training time and test time. GS improved results without ST, but had little effect with ST. 

\begin{table}[H]
\centering
\footnotesize
\begin{tabular}{lcc} \hline\hline
Model               & Accuracy   \\ \hline
\textsc{+GS +ST}          & 84.0   \\
\textsc{+GS -ST}          & 61.0   \\
\textsc{-GS +ST}          & 84.7    \\
\textsc{-GS -ST}          & 54.7   \\
\hline
\etoebaseline{} & 86.5 \\
\hline\hline
\end{tabular}
\caption{Results of \etoebaseline{} compared to our model (with GS and ST variants) on hypernym extraction.}
\label{tab:toy_results}
\end{table}